%% file: acl2023.tex
\pdfoutput=1

\documentclass[11pt]{article}

\usepackage[final]{ACL2023}

\usepackage{times}
\usepackage{latexsym}
\usepackage{booktabs}
\usepackage{hyperref}
\usepackage{graphicx}
\usepackage{multirow}
\usepackage{arydshln}
\usepackage{svg}
\usepackage{caption}
\usepackage{tablefootnote}
\usepackage{booktabs}
\usepackage{inconsolata}
\usepackage[T1]{fontenc}

\usepackage[utf8]{inputenc}

\usepackage{microtype}

\usepackage{inconsolata}

\title{Automatic Data Retrieval for Cross Lingual Summarization}

\author{\textbf{Nikhilesh Bhatnagar}\textsuperscript{1}\textsuperscript{*} \enspace  \textbf{Ashok Urlana}\textsuperscript{2}\textsuperscript{*}\enspace \textbf{Vandan Mujadia}\textsuperscript{1} \\
\enspace \textbf{Pruthwik Mishra}\textsuperscript{1} \enspace  \textbf{Dipti Misra Sharma}\textsuperscript{1}\\
IIIT Hyderabad\textsuperscript{1} \enspace \enspace \enspace \enspace \enspace \enspace TCS Research, Hyderabad, India\textsuperscript{2}
\\
{\tt tingc9@gmail.com}, {\tt ashok.urlana@tcs.com}, {\tt vandan.mu@research.iiit.ac.in} \\ {\tt pruthwik.mishra@research.iiit.ac.in}, {\tt dipti.m@iiit.ac.in}
}

\begin{document}
\maketitle
\def\thefootnote{*}\footnotetext{Equal Contribution}
\begin{abstract}
Cross-lingual summarization involves the summarization of text written in one language to a different one. There is a body of research addressing cross-lingual summarization from English to other European languages. In this work, we aim to perform cross-lingual summarization from English to Hindi. We propose pairing up the coverage of newsworthy events in textual and video format can prove to be helpful for data acquisition for cross lingual summarization. We analyze the data and propose methods to match articles to video descriptions that serve as document and summary pairs. We also outline filtering methods over reasonable thresholds to ensure the correctness of the summaries. Further, we make available 28,583 mono and cross-lingual article-summary pairs\footnote{\url{https://github.com/tingc9/Cross-Sum-News-Aligned}}. We also build and analyze multiple baselines on the collected data and report error analysis.
\end{abstract}

\section{Introduction}

In the field of Natural Language Processing (NLP), advancements in human language understanding and processing have often required extensive data. Cross-lingual summarization, a recent focus in NLP, has seen the emergence of well-known datasets like CrossSumm \cite{bhattacharjee2021crosssum}  and PMIndiaSum \cite{urlana2023pmindiasum}. In the context of Indian languages, recent datasets such as XL-Sum \cite{hasan-etal-2021-xl}, MassiveSumm \cite{varab-schluter-2021-massivesumm}, and PMIndiaSum \cite{urlana2023pmindiasum} promote research in low-resource language setting.

Generating human-annotated datasets is a labor-intensive and costly endeavor \cite{urlana-etal-2022-tesum}. To address the scarcity of resources for low-resource languages, some efforts involve scraping news websites and use source document prefixes or headlines as summaries. However, this approach has two main shortcomings: 1) It often lacks diversity in information since all article-summary pairs originate from a single domain, and 2) Many source documents lack comprehensive coverage of specific events or incidents.



In this work, we present the data collection and alignment approach aimed at enhancing both mono and cross-lingual summarization for Indian languages. Our core hypothesis is that news events can serve as a basis for aligning diverse sources of coverage in the context of summarization. Specifically, we concentrate on textual summarization and propose that by pairing up YouTube descriptions with corresponding news article coverage of different events from a multitude of sources, we can create a substantial summarization dataset comprising of 28,583 pairs. We have experimented with state-of-the-art multilingual summarization models and performed error analysis to assess the effectiveness of the pretrained models for both mono and cross-lingual summarization. 

\begin{figure*}[ht]
    \centering 
    \includegraphics[width=\textwidth]{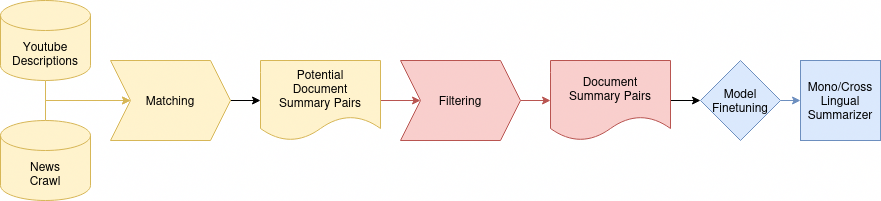}
    \caption{Overall architecture}
    \label{architecture}
\end{figure*}

In the subsequent sections, we outline related work, our data collection methodology, the process of matching summarization pairs, and the subsequent filtering to derive the final dataset for our work in \ref{architecture}. Furthermore, we describe the training of various baseline models using the collected data and present our research findings.

\begin{table}
\centering
\begin{tabular}{llll} 
\toprule
                        & \textbf{En-En} & \textbf{En-Hi} & \textbf{Hi-Hi}  \\ 
\midrule
Unfiltered pairs\tablefootnote{Only pairs with >= 0.5 similarity are selected}        & 80868          & 210056         & 276599          \\ 
\hline
Article  Summary        & 78183          & 204975         & 274342          \\
Deduplication           & 49116          & 133538         & 148421          \\
Unigram Overlap\tablefootnote{0.4 for mono-lingual summarization and 0.3 for cross-lingual summarization}         & 25898          & 33054          & 68025           \\
Cosine Similarity\tablefootnote{Pairs above 0.7 threshold were selected} & 6340           & 7892           & 14351           \\ 
\hline
\textbf{Filtered pairs}          & \textbf{6340}  & \textbf{7892}  & \textbf{14351}  \\
\bottomrule
\end{tabular}
\caption{Data filtering and statistics }
\label{tab:data_statistics}
\end{table}

\begin{table*}[t!]
    \centering
    \begin{tabular}{ll}
    \includegraphics[width=\columnwidth]{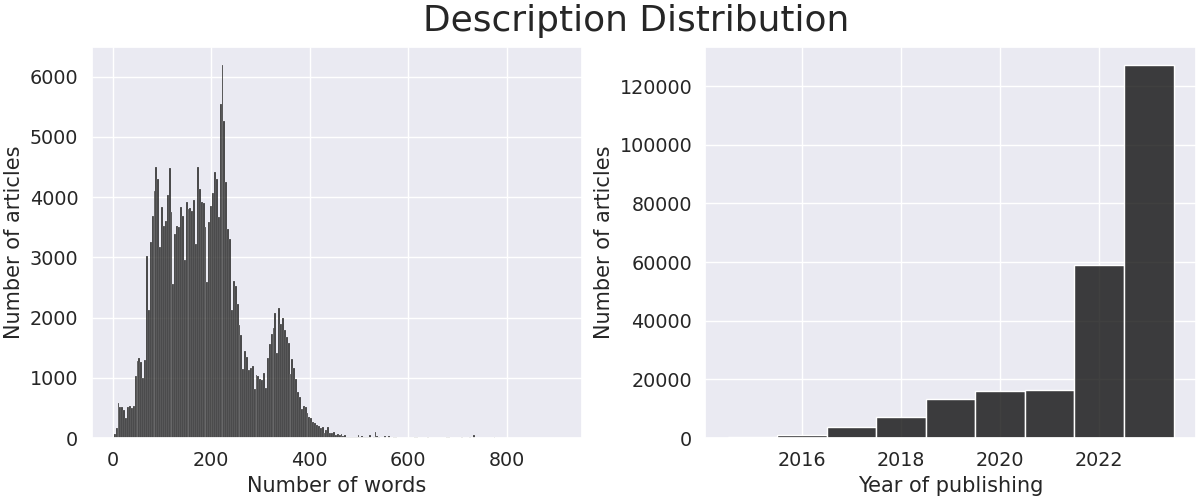}

    \includegraphics[width=\columnwidth]{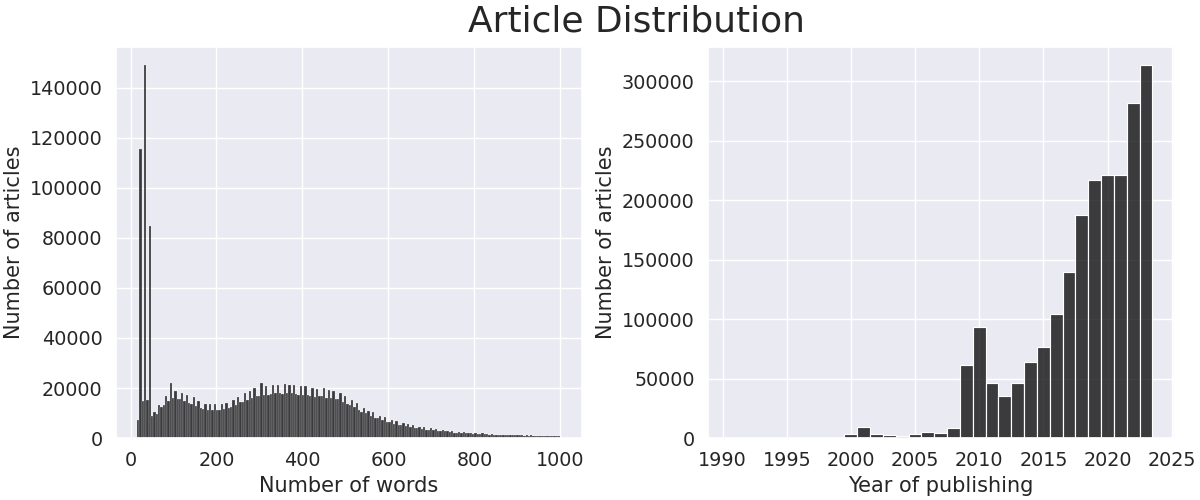}
    \end{tabular}
    
    \label{tab:my_label}
\end{table*}

\section{Literature Survey}

Significant progress has been made in the field of English language summarization, in stark contrast to the relatively limited efforts focused on Indian languages in the context of summarization and related Natural Language Generation (NLG) tasks \cite{urlana-etal-2022-tesum, urlana2022indian, urlana2023pmindiasum}, including headline generation. Nevertheless, recent times have witnessed a surge in active research in this domain, notably marked by the release of datasets like XL-Sum \cite{hasan-etal-2021-xl}, MassiveSumm \cite{varab-schluter-2021-massivesumm}, and others. These multilingual datasets comprise pairs of articles and summaries extracted from publicly available news sources, encompassing a range of Indian languages, including Hindi, Gujarati, Bengali, and more.

Furthermore, the IndicNLG Suite \cite{kumar-etal-2022-indicnlg} has contributed by providing datasets designed for various Indic language NLG tasks, such as sentence summarization and headline generation. Despite these promising developments, there is still a need for continued effort in this area to attain performance levels in Indian language summarization that is on par with their English language counterpart.


\section{Data Creation}
There are three steps for data preparation - Data Collection, Pair Matching and lastly, Pair Filtering.
\subsection{Data Collection}
For the potential summaries, we crawl the language specific youtube channels of 4 sources - NDTV, IndianExpress, OneIndia and TheQuint. In this work, we focus on English and Hindi languages as a proof of concept. We used yt-dlp to crawl the descriptions of the above mentioned channels and populate our description store. For the potential documents, we crawl the respective language websites for each of the outlined sources using news-please \cite{Hamborg2017}. In total, we crawled ~150K youtube descriptions and ~350K news articles. We preprocess and compute the distUSE \cite{DBLP:journals/corr/abs-1907-04307} vectors for the text and the extracted title of the respective articles and descriptions.


\subsection{Pair Matching}
\begin{figure*}[tp]
    \centering 
    \includegraphics[width=1\textwidth]{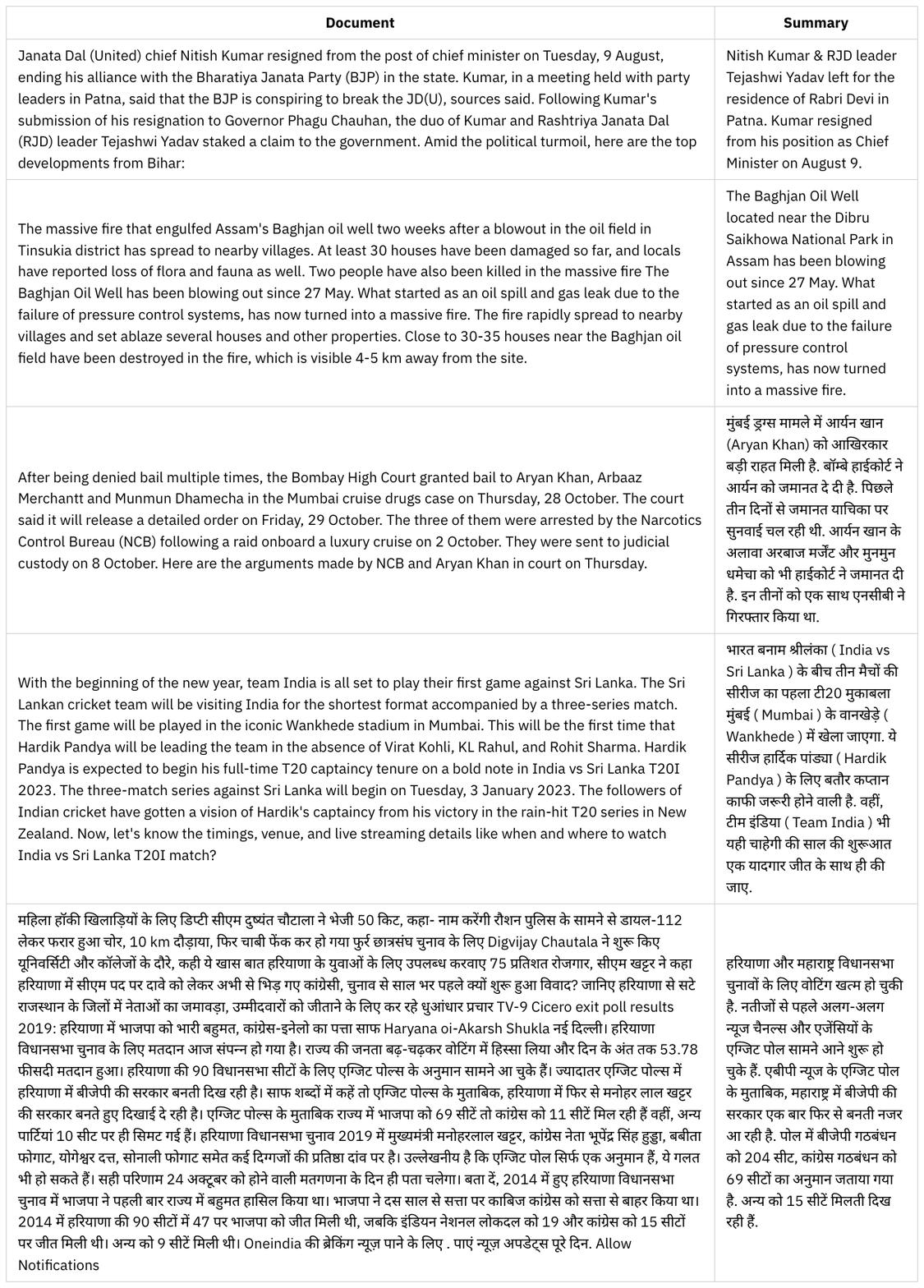}
    \caption{Summary examples}
    \label{examples}
\end{figure*}

The aggregated data needs to be aligned, resulting in pairs of articles and descriptions that include multiple pairs of document-summaries for a single event. A quick analysis of the data reveals that for the purposes of abstractive summarization, the fraction of common content words in the summary and the document must be high and the two texts must be semantically similar. We have three factors driving the pair matching \ref{tab:data_statistics}
\begin{enumerate}
    \item Date of incidence: The article and description must be published within 2.5 days of each other.
    \item Unigram Overlap: The fraction of non stop word unigrams in the summary that are present in the document. For the cross-lingual case we perform a parallel dictionary lookup or transliterate, if the lookup fails.
    \item Semantic Similarity: The cosine similarity of the DistUSE \cite{DBLP:journals/corr/abs-1907-04307} text embeddings for the two texts.
\end{enumerate}
Some samples are presented in Figure~\ref{examples}.

\subsection{Pair Filtering}

We overgenerate the pairs based on the above factors and employ the following mechanisms to filter out bad summaries.
\begin{enumerate}
    \item Preprocessing: Deduplication and length-based outlier removal.
    \item Compression Ratio: Summaries within one s.d. of the compression ratio are kept.
    \item Unigram Overlap: Summaries with more than 0.4 for monolingual and 0.3 overlap for crosslingual case are kept.
    \item Similarity Threshold: Only pairs with both title and text similarity above 0.7 are kept.
\end{enumerate}

After the whole process, we are left with ~8K article summary pairs for cross lingual summarization and ~14K pairs for monolingual summarization. We release the combined data of 28,583 article-summary pairs for en-en, en-hi and hi-hi language pairs.

\section{Methodology}

Our objective is to establish a benchmark for evaluating traditional summarization methods. Below, we present the following paradigms and specify the language settings utilized for testing with each approach.

\subsection{Training free-baselines:}

\textbf{Extractive Method:}  In this approach, we incorporate two training free-baselines:
   \begin{itemize}
       \item Selection of the lead sentence.
       \item Scoring each sentence in the document against a reference and choosing the best one in an oracle-like manner.
   \end{itemize}

\input{training_free_baselines}

\input{data_splits}

\input{multi_lingual_baselines}
\input{error_analysis}

\subsection{Pre-trained models}
\begin{itemize}
    \item \textbf{IndicBART} \cite{dabre2021indicbart} is a sequence-to-sequence model (244M parameters) trained on 11 Indic languages alongside English. Its training approach follows a masked span reconstruction objective, similar to MBart.

    \item \textbf{mBART:} BART \cite{lewis2019bart} is a denoising auto-encoder, the multilingual version mBART \cite{chipman2022mbart} (610M parameters) designed for the pretraining of sequence-to-sequence models.
    
    \item \textbf{mT5:} The T5 \cite{raffel2020exploring} employs an encoder-decoder Transformer architecture fine-tuned on the C4 corpus. We use the multilingual mT5 \cite{xue-etal-2021-mt5} model for finetuning on our data.
\end{itemize}
\input{experimental_setup}

\section{Experiments and results}
\noindent We fine-tune on three pretrained models namely IndicBART, mBART and mT5-base. The dataset splits are detailed in the Table~\ref{tab:data_splits}. We have used 80\% for the training purpose and 10\% each for the validation and testing. We report the average ROUGE\cite{lin2004rouge} f1 scores in \ref{tab:multi_lingual_baselines}. We can see from Table~\ref{tab:training_free_baselines}, that for en-en samples, more than half the samples the summary is the same as the first sentence of the document. As mentioned in the Table~\ref{tab:multi_lingual_baselines}, for all the three language pairs mBART-large50 model outperforms the remaining two models (IndicBART, mT5). Compared to IndicBART, mBART outperforms.
\subsection{Error analysis}
We performed the error analysis by following the guidelines mentioned in the PMIndiaSum \cite{urlana2023pmindiasum}. In case of English and Hindi mono-lingual summarization, IndicBART makes lot of grammatical errors by omitting the relevant information. For cross-lingual (En-Hi) summarization, we have observed that all three models make faithfulness errors by omitting the relevant information. Overall, we find that mBART performs better in case of both mono and cross-lingual summarization methods.

\section{Conclusions}
In this study, we have developed a mono and cross-lingual summarization dataset for English-Hindi language pairs by leveraging a variety of news events and their corresponding YouTube descriptions. Our experimentation with multilingual models revealed that mBART consistently outperforms the IndicBART and mT5 models. Our error analysis indicates significant opportunities for the development of more effective multilingual models for low-resource languages.
\section{Limitations}
The dataset is scraped from various news sources, there is a possibility of code-mixed samples in the document-summary pairs. Moreover, because of slight differences in the content of each source, not all summaries are equal.
\section{Future Work}
It is likely that for other Indian Languages, the collected data from this method would be a fraction so it would be worth looking into training a single model for cross lingual summarization where the parameter sharing could benefit low resource languages.

\bibliography{custom, anthology}
\bibliographystyle{acl_natbib}

\end{document}

%% file: training_free_baselines.tex
\begin{table}[htb]
\centering\small
\begin{tabular}{cccclcc}
\toprule
 & \multicolumn{3}{c}{\textbf{LEAD}} & \multicolumn{3}{l}{\textbf{EXT-ORACLE}} \\ 

 & \multicolumn{1}{l}{R-1} & \multicolumn{1}{l}{R-2} & \multicolumn{1}{l}{R-L} & \multicolumn{1}{l}{R-1} & \multicolumn{1}{l}{R-2} & \multicolumn{1}{l}{R-L} \\ 
\cmidrule(lr){2-4}\cmidrule(lr){5-7}
\textit{\textbf{en-en}} & \multicolumn{1}{c}{43.2} & \multicolumn{1}{c}{26.3} & 34.9 & \multicolumn{1}{c}{50.1} & \multicolumn{1}{c}{33.1} & 41.7 \\ 
\textit{\textbf{hi-hi}} & \multicolumn{1}{c}{10.4} & \multicolumn{1}{c}{4.1} & 9.6 & \multicolumn{1}{c}{24} & \multicolumn{1}{c}{11.3} & 21.9 \\ \bottomrule
\end{tabular}
\caption{ROUGE scores of training free-baselines}
\label{tab:training_free_baselines}
\end{table}

%% file: data_splits.tex
\begin{table}[htb]
\centering
\begin{tabular}{cccc}
\toprule
\textbf{Lang-pair} & \textbf{Train} & \textbf{Valid} & \textbf{Test} \\ \midrule
\textit{\textbf{en-en}} & 5072 & 634 & 634 \\ 
\textit{\textbf{en-hi}} & 6314 & 789 & 789 \\ 
\textit{\textbf{hi-hi}} & 11482 & 1435 & 1435 \\ \bottomrule
\end{tabular}%
\caption{Data splits counts}
\label{tab:data_splits}
\end{table}


%% file: multi_lingual_baselines.tex
\begin{table*}[h!t!]
\centering
\begin{tabular}{cccccccccc}
\toprule
 & \multicolumn{3}{c}{\textbf{IndicBART}} & \multicolumn{3}{c}{\textbf{mBART}} & \multicolumn{3}{c}{\textbf{mT5-base}} \\ 
 \cmidrule(lr){2-4}\cmidrule(lr){5-7}\cmidrule(lr){8-10}
 & \multicolumn{1}{c}{R-1} & \multicolumn{1}{c}{R-2} & R-L & \multicolumn{1}{c}{R-1} & \multicolumn{1}{c}{R-2} & R-L & \multicolumn{1}{c}{R-1} & \multicolumn{1}{c}{R-2} & R-L \\ 
 \cmidrule(lr){2-4}\cmidrule(lr){5-7}\cmidrule(lr){8-10}
\textit{\textbf{en-en}} & \multicolumn{1}{c}{46.7} & \multicolumn{1}{c}{28.1} & 36.8 & \multicolumn{1}{c}{\textbf{49}} & \multicolumn{1}{c}{\textbf{31}} & \textbf{40.1} & \multicolumn{1}{c}{42.4} & \multicolumn{1}{c}{24.9} & 34.1 \\ 
\textit{\textbf{en-hi}} & \multicolumn{1}{c}{20} & \multicolumn{1}{c}{7.8} & 14.8 & \multicolumn{1}{c}{\textbf{25.7}} & \multicolumn{1}{c}{\textbf{10.7}} & \textbf{17.6} & \multicolumn{1}{c}{18} & \multicolumn{1}{c}{5.1} & 13.1 \\ 
\textit{\textbf{hi-hi}} & \multicolumn{1}{c}{28.2} & \multicolumn{1}{c}{15} & 21.5 & \multicolumn{1}{c}{\textbf{40.7}} & \multicolumn{1}{c}{\textbf{26}} & \textbf{32.2} & \multicolumn{1}{c}{30.6} & \multicolumn{1}{c}{15} & 22.3 \\ \bottomrule
\end{tabular}%
\caption{Multi-lingual baselines ROUGE scores}
\label{tab:multi_lingual_baselines}
\end{table*}

%% file: error_analysis.tex
\begin{table*}[h!t!]
\centering
\resizebox{\textwidth}{!}{%
\begin{tabular}{cccccccccc}
\toprule
\multirow{2}{*}{\textbf{Error Types}} &  \multicolumn{3}{c}{\textbf{en-en}} &  \multicolumn{3}{c}{\textbf{hi-hi}} & \multicolumn{3}{c}{\textbf{en-hi}} \\ 
\cmidrule(lr){2-4}\cmidrule(lr){5-7}\cmidrule(lr){8-10}
 & \multicolumn{1}{c}{IndicBART} &  \multicolumn{1}{c}{mBART} & mT5-base &  \multicolumn{1}{c}{IndicBART} & \multicolumn{1}{c}{mBART} &  mT5-base & 
 \multicolumn{1}{c}{IndicBART} & \multicolumn{1}{c}{mBART} &  mT5-base \\ 
 \midrule
\textbf{Comprehensibility} & \multicolumn{1}{c}{0} & \multicolumn{1}{c}{0} & 0 & \multicolumn{1}{c}{3} & \multicolumn{1}{c}{1} & 0 & \multicolumn{1}{c}{12} & \multicolumn{1}{c}{0} & 1 \\ 
\textbf{Grammar} & \multicolumn{1}{c}{29} & \multicolumn{1}{c}{2} & 0 & \multicolumn{1}{c}{21} & \multicolumn{1}{c}{14} & 19 & \multicolumn{1}{c}{2} & \multicolumn{1}{c}{1} & 0 \\ 
\textbf{Factuality} & \multicolumn{1}{c}{6} & \multicolumn{1}{c}{6} & 6 & \multicolumn{1}{c}{10} & \multicolumn{1}{c}{11} & 12 & \multicolumn{1}{c}{20} & \multicolumn{1}{c}{36} & 28 \\ 
\textbf{Omission} & \multicolumn{1}{c}{15} & \multicolumn{1}{c}{18} & 29 & \multicolumn{1}{c}{20} & \multicolumn{1}{c}{11} & 8 & \multicolumn{1}{c}{20} & \multicolumn{1}{c}{23} & 12 \\ 
\textbf{Redundancy} & \multicolumn{1}{c}{10} & \multicolumn{1}{c}{13} & 6 & \multicolumn{1}{c}{7} & \multicolumn{1}{c}{5} & 17 & \multicolumn{1}{c}{17} & \multicolumn{1}{c}{3} & 34 \\ 
\hdashline
\textbf{No error} & \multicolumn{1}{c}{8} & \multicolumn{1}{c}{22} & 16 & \multicolumn{1}{c}{4} & \multicolumn{1}{c}{16} & 6 & \multicolumn{1}{c}{0} & \multicolumn{1}{c}{1} & 0 \\ \bottomrule
\end{tabular}%
}
\caption{Error analysis on different models and various language combinations}
\label{tab:error_analysis}
\end{table*}

%% file: experimental_setup.tex
\begin{table}[h]
\center
\setlength{\tabcolsep}{0.25ex}
\begin{tabular}{lrrr}
\toprule
\textbf{Parameters} & \textbf{mBART} & \textbf{mT5} & \textbf{IndicBART} \\ \midrule
Max source length & 512   & 512   & 512   \\ 
Max target length & 128    & 128  & 128    \\ 
Batch Size        & 2     & 1     & 4     \\ 
Epochs            & 5     & 5     & 20    \\ 
Vocab Size        & 50265 & 32128 & 64015 \\ 
Beam Size         & 4     & 4     & 4     \\ 
Learning Rate     & 5e-5  & 5e-5      & 5e-5   \\ \bottomrule
\end{tabular}
\caption{Experimental setup and parameters settings}
\label{tab:parameters}
\end{table}